# Autonomous surveillance for biosecurity

Raja Jurdak[1], Alberto Elfes[1], Branislav Kusy[1], Ashley Tews[1], Wen Hu[1,2], Emili Hernandez[1], Navinda Kottege[1], and Pavan Sikka[1]

[1]CSIRO Autonomous Systems, 1 Technology Court, Pullenvale, QLD 4069, Australia

[2]University of New South Wales, School of Computer Science and Engineering, Sydney, NSW 2052, Australia

*Corresponding author:* Jurdak, R. (rjurdak@ieee.org).

**The global movement of people and goods has increased the risk of biosecurity threats and their potential to incur large economic, social, and environmental costs. Conventional manual biosecurity surveillance methods are limited by their scalability in space and time. This article focuses on autonomous surveillance systems, comprising sensor networks, robots, and intelligent algorithms, and their applicability to biosecurity threats. We discuss the spatial and temporal attributes of autonomous surveillance technologies and map them to three broad categories of biosecurity threat: (i) vector-borne diseases; (ii) plant pests; and (iii) aquatic pests. Our discussion reveals a broad range of opportunities to serve biosecurity needs through autonomous surveillance.**

**Why autonomous surveillance for biosecurity?**

The ease of movement of people and goods coupled with rapid development have significantly increased the risk of biological threats to fragile ecosystems and economic activity across the world [1–3], with diseases, pests, and invasive species diffusing more easily across regions and borders. These threats incur significant social, economic, and environmental costs [4–6]. Social costs arise from vector-borne diseases that impact humans. Economic costs are for the time and effort involved in implementing detection and control strategies, particularly for invasive species of plants, animals, and insects in primary production. The environmental cost arises from the detrimental effect the threats have on the environment and its ability to cope and adapt. Conventional early detection systems are fundamentally undertaken by human surveillance of valuable asset lands or, increasingly, by integration of fixed sensor nodes into the detect–react–manage cycle. Current systems, however, do not meet key biosecurity needs as they are labor intensive and costly, they cannot cover environments that pose risks to humans, and they are limited in spatial coverage.

The global scale of movement of biosecurity threats has also created a need for agricultural producers to show compliance with regulatory and market access requirements. All of these drivers have increased interest in autonomous surveillance systems to prevent, detect, and manage biosecurity threats [7–9]. Sensing and robotic systems have recently been deployed to track flying foxes as disease vectors [10], to detect fruit flies [11], to differentiate weeds from healthy plants [12], and to conduct tropical forest surveys to detect *Miconia* invasions [13].

Despite significant progress in autonomous surveillance technology, its application for biosecurity presents several new challenges. The first challenge is around spatial scale and resolution. Biosecurity threats typically span entire regions, countries, or even continents. Detection technologies therefore need to be distributed over large areas and provide sufficient spatial resolution to capture any threats. For instance, the spread of vector-borne diseases [14] such as Hendra virus requires monitoring of the originating vectors (fruit bats) at the continental scale of their movement [10], which can span several countries. Weed detection, by contrast, requires high spatial resolution rather than wide spatial coverage, through hyperspectral imaging for targeted pesticide treatment [15] to avoid blanket spraying and pesticide resistance.

The second challenge for autonomous surveillance in biosecurity is duration and temporal resolution. Biosecurity threat detection and management requires continuous monitoring of vulnerable regions and assets, posing significant challenges for the design of autonomous surveillance systems, mainly due to the limited energy supply to power these systems in the field. The purpose of autonomous surveillance is to prevent the introduction of a threat to a previously unaffected area (for instance, preventing foot-and-mouth disease from reaching Australian cattle [16]) or to limit the expansion of an existing threat, such as limiting the population numbers of fruit flies [17].

The final challenge for autonomous surveillance in biosecurity is ensuring fitness of purpose. Biosecurity threats are often multidimensional [2,3,9], with diverse monitoring and management requirements depending on context. The temporal and spatial scale at which a threat propagates in the landscape varies from slow in the case of invasive weeds growing in a rainforest [13] to fast in the case of disease outbreaks [18]. This diversity of how quickly biosecurity problems spread requires matching technology solutions [19]. Tailoring the technology solution to address the multiple dimensions of threats is challenging. To keep an area pest free through continuous monitoring is best served through a network of fixed sensors. However, the optimal locations of the fixed sensors could be determined through an initial survey of the area using a robotic platform. Once pest incursions are detected, the system should take action using the fixed sensor nodes themselves or a separate robotic platform. The integration and cooperation between existing platforms presents a clear challenge and opportunity to serve biosecurity needs.

In light of the above challenges, we present recent developments in autonomous surveillance for biosecurity. The discussion maps autonomous surveillance solutions to three broad biosecurity areas (Table 1): (i) vector-borne diseases; (ii) plant pests; and (iii) aquatic pests. We discuss the spatial and temporal sampling features of each technology separately and then analyze their fitness for the three biosecurity areas and for specific applications within those areas. The detailed mapping of autonomous surveillance technologies to biosecurity problems reveals a broad range of potential application areas of opportunity, ranging from tracking of disease-borne vertebrates, insects, or cattle, through distributed sensing technology and mobility modelling and the detection and management of plant pests through stationary and mobile sensors and aerial and ground robots, to the detection of aquatic pests through combinations of autonomous underwater vehicles (AUVs) and grids of bioacoustics monitoring nodes.

**Table 1. Summary of autonomous surveillance technologies, their spatial and temporal attributes, and their applicability to specific biosecurity areas**

| Technology | Duration | Temporal resolution | Spatial resolution | Spatial coverage | Adaptivity | Utility | Biosecurity application area |
|---|---|---|---|---|---|---|---|
| *Fixed sensors* | Long | High | Low | Low | Temporal | Continuous monitoring /communications relays | Vector-borne disease, plant pests, aquatic pests |
| *Mobile sensors* | Medium | High | Low | High | Temporal, partial spatial | Individual-based monitoring | Vector-borne disease, aquatic pests |
| *Ground robots* | Short | High | High | Low | Temporal and spatial | Terrestrial surveys/management | Plant pests |
| *Aquatic robots* | Variable | High | High | Medium | Temporal and spatial | Aquatic surveys/management | Aquatic pests |
| *Aerial robots* | Variable | High | High | High | Temporal and spatial | Aerial surveys/management | Plant/aquatic pests |

**Sensors**

*Fixed sensors*

Over the past decade, wireless sensor networks (WSNs), which comprise numerous different sensors such as microclimate and multimedia sensors connected wirelessly over radio, have been deployed successfully in the field to detect biosecurity threats such as foot-and-mouth disease [16], insect pests [20], feral animals [21], invasive frogs [22] and fish [23]. Compared with other approaches, WSNs provide high temporal frequency observation and can operate independently for a long period of time, but have limited spatial resolution and small spatial coverage because of the infrastructure deployment and maintenance cost constraints. Therefore, they can provide effective methods for small-scale continuous monitoring applications. Furthermore, they can be used as communication relays to cloud services because many of their nodes are equipped with cellular network communication (e.g., 3G, 4G) components.

Biosecurity WSNs typically collect audio, image, and body temperature observations produced by monitored targets and use machine learning and pattern recognition algorithms to identify targets of interest automatically from these observations. For example, Rainwater-Lovett *et al.* used infrared cameras to collect the body temperature of cattle, which can then be used to infer whether the cattle have been infected by foot-and-mouth disease [16], as higher body temperatures are indicative of this disease.

Low-power image sensor networks have been used to detect and classify insect pests [20] and feral animals [21]. An autonomous insect monitoring (AIM) device was developed that was capable of detecting and classifying insect pests in the field [11]. In this design, an open tunnel is inserted to slow the motion of insects and a sensor network camera node takes a picture of the insects as they move through the camera chamber. The pictures of insects are then analyzed automatically to count the number of insects as well as to classify them.

Bioacoustic sensor networks have been developed and deployed to detect invasive frogs (cane toads) [22] and invasive fish [23] from their calls. For example, *Tilapia mariae* is native to West African coastal drainages in the Gulf of Guinea and became naturalized in the USA and Australia due to aquarium and aquaculture releases. *T. mariae* is a declared noxious fish in Australia due to a potential detrimental impact on native species. Kottege *et al.* developed detection and classification methods for short-duration broadband sounds, which can be used to detect the presence of *T. mariae* in large bodies of fresh water [23].

*Mobile sensors*

One of the main limitations of fixed sensors is their relatively small spatial coverage of the environment. Motion detectors are limited to a range of a few tens of meters and multimedia sensors can cover areas of at most a few hundred meters due to the limited resolution of low-power hardware [24]. Attaching sensors to mobile objects such as domestic and wild animals has the potential to greatly extend the spatial coverage of fixed sensors.

Recent growth in personal mobile computing has lead to improvements in the size and energy efficiency of electronic devices, enabling their deployment on small-sized animals. Dubbed 'One Giant Leap for Wildlife Tracking' [25], lightweight telemetric tags were deployed on hummingbirds [26], pigeons [27], toucans [28], and flying foxes [10]. Furthermore, miniature inertial measurement units, microphones, or weather sensors can be used to classify animal activity and context, such as urination or defecation [16], which can indicate the shedding of a virus, seed dispersal or energetics [10], which are relevant for predictive movement modeling of vectors. While satellite trackers remain quite bulky with a minimal weight of about 5 g [25], smaller radio frequency identification devices (RFIDs) weighing 0.2–1 g have been used to track insects such as bees, beetles, and dragonflies [29].

The ability to track the location and activity of individual animals can significantly contribute to our understanding of disease vector pathways and the behavioral patterns of aquatic, airborne, and land pests. The technology can provide data in near real time, which allows dissemination of timely biosecurity alerts in the affected areas. Spatiotemporal movement data are also an important step toward developing accurate behavioral models [30] of the animal species under study. Agent-based modeling and simulations of animal populations [31] have applications in prediction of the disease risks that the animals carry and the management of pest animal species, which we discuss in more detail below.

**Robots**

Robots are used to acquire data at high spatiotemporal resolution as well as to provide quantitative and qualitative analysis of the data. They comprise an onboard computer, internal sensing of their state, external sensing of the environment, and some form of mobility in terrestrial (Box 1), aerial (Box 2), or aquatic (Box 3) terrains. They are capable of traversing and sampling the environment at predetermined times and locations, on an event trigger from the static sensor network, by user request, or autonomously. Robots have higher processing capability and carry a higher-quality sensor payload than fixed nodes, such as high-resolution color, thermal, or hyperspectral cameras [32], biomass, soil, and atmospheric sensors, and 3D ranging systems useful for modeling or mapping areas of interest [33]. Given that they are mobile, they can interrogate different areas of the environment in greater detail and conduct onboard processing and reasoning, thereby reducing communication throughput and providing analytical results and state information independently or through the same network as the static nodes. They can also interact with the environment by picking and analyzing samples [34], reducing the requirement for human presence. A team of robots has the potential to be operated by a single human.

**Biosecurity applications of autonomous surveillance**

Autonomous surveillance has clearly emerged as a significant tool to address biosecurity needs, although its full potential for managing threats remains to be realized. In particular, there remains an open challenge in the biosecurity of detecting and localizing threats and targeting interventions to the specific time/location of the threat. For instance, targeted interventions for vector-borne diseases involve localizing regions of high risk and focusing available resources on managing the risk in these regions. Similarly, weed eradication needs targeted spraying of only weed-infested patches within an agricultural field to avoid the development of pesticide resistance, which has become a major issue particularly in the USA, Europe, and Australia.

To localize threats and target interventions requires systems that operate at both high spatial and high temporal resolution, for a sufficiently long duration, and over a large enough area. No single autonomous surveillance technology (Table 1) meets all of those needs, but, collectively, these technologies can achieve the spatial and temporal performance necessary for targeted interventions.

Consider the surveillance of vector-borne diseases such as mosquito-borne dengue or chikungunya or fruit-bat-borne Ebola or Hendra virus [35]. Current technology solutions range from position and activity tracking of individual vectors using dedicated devices [36] to biomass estimation of the vectors at a macroscale by surveys or traps [37]. Estimation of vector prevalence provides only coarse-grained spatial and temporal detection of threats, while individual-based tracking captures the fine-grained behavior of individuals, such as activity, state of health, and contact with other individuals, and can be used to model their responses to various social, temporal, spatial, and environmental contexts. These data can then drive predictive models of biosecurity-related risks such as transmission of diseases or crop damage and their distribution within a landscape. Predictive modeling is a complex task that requires an understanding of how the disease vectors (e.g., flying foxes) respond, in terms of their movement and choice of migration routes and foraging locations, to the structure of landscapes and the distribution of resources within them and how this varies across landscapes, seasons, and individuals [30, 31]. A common approach is to develop agent-based models characterizing the behavior of individual animals based on empirical spatiotemporal data from telemetry tags. The agent-based modeling framework is a powerful tool that allows us to study biosecurity-related risks in a specific spatial, temporal, environmental, and social context by diffusing the risk across the landscape based on the simulated behavior of individual disease vectors and their interaction with the environment [18]. Based on risk estimates, ground or aerial robotic platforms can be dispatched into high-risk regions to investigate further or take action to limit or control the spread of the disease.

For plant pests, cooperation among multiple layers of autonomous surveillance technologies is equally important. Weeds, disease spores, and insects that damage plants can be addressed through pesticides; however, blanket administration of pesticides has led to the development of high pesticide resistance in many countries. It is imperative, therefore, to localize the pests in space and time for targeted pesticide spraying. Fixed sensors are best suited to monitor areas of interest at regular intervals to detect the likelihood of plant pests and once these are detected they can localize the threat and notify a ground or aerial robot platform to move into the region for more detailed inspection and actuation to eliminate the threat; for instance, through selective application of pesticide. This closed-loop system combining sensing, analyzing, and acting on the threat will require a complex analysis of the threat that is application specific. As an example, preventing fruit fly incursions would require real-time image classification at automated insect traps to identify the presence of the species of interest, while weed detection requires capture and classification of hyperspectral images to differentiate weeds from healthy plants, to direct autonomous robots to where to intervene.

A similar logic applies to aquatic pest surveillance. Aquatic robots are the primary platform for applications such as port surveillance and marine surveys to protect against importation of invasive species, such as the Asian green mussel into Australia[1]. However, how often should surveys be conducted and where should they focus in detecting threats remain open questions. The inclusion of fixed aquatic sensors to collect regular samples, to trigger aquatic robots to initiate a survey, and to guide these robots to areas of high threat likelihood can close the loop and combine the long-term operation of fixed sensors with the high spatial and temporal resolution of aquatic robots. For instance, a combination of 50 floating wireless sensors was used to measure the temperature in a lake at six different depths to trigger an autonomous surface vessel (ASV) to investigate possible algal blooms [38].

---

[1] [Department of Fisheries, Western Australia (2014) *Asian Green Mussel Factsheet* (http://www.fish.wa.gov.au/Documents/biosecurity/marine_pest_fact_sheet_asian_green_mussel.pdf)]

While the current state of the art in autonomous surveillance technology is useful for many biosecurity threats, certain threats will spread quickly and will require further advancements. Swarms of locusts, for instance, may spread at a faster spatiotemporal scale than current technology can cover. Such fast-moving threats will require multiple robotic platforms for simultaneous coverage of the areas in question at high spatial resolution, possibly complemented by fixed sensors for continuous sampling and event notifications at key points. For instance, the control of two unmanned aerial vehicles (UAVs) was coordinated by speed modulation to synchronize their flights for the aerial detection of the fungus-like organism *Phytophthora infestans* [39]. While this work serves as an initial proof of concept for the use of multiple robots to detect biosecurity threats, it still deals with a relatively slow-moving threat of fungus spread. Using multiple robots simultaneously to detect biosecurity threats is an area still in its infancy with immense opportunities.

**Concluding remarks and future perspectives**

With the maturation of autonomous surveillance technologies, we expect that these systems will increasingly be adopted for targeted detection, localization, and management of biosecurity threats. Closed-loop solutions that combine multiple layers of autonomous surveillance technologies for optimal threat management will provide the highest return on investment. Given the diversity of biosecurity threats, we expect that the development of smart algorithms to drive this sense–think–act loop will be a highly active research area in the next decade.

53 Ribas, D. *et al.* (2012) Girona 500 AUV: from survey to intervention. *IEEE ASME Trans. Mech.* 17, 46–53

54 Singh, H. *et al.* (2004) Seabed AUV offers new platform for high-resolution imaging. *Eos Trans. Am. Geophys. Union* 85, 289–296

55 Thompson, D. *et al.* (2013) *MBARI Dorado AUV's Scientific Results*, IEEE

56 Sharp, K.M. and White, R.H. (2008) *More Tools in the Toolbox: The Naval Oceanographic Office's Remote Environmental Monitoring UnitS (REMUS) 6000 AUV*, IEEE

57 Dunbabin, M. *et al.* (2005) *A Hybrid AUV Design for Shallow Water Reef Navigation*, IEEE

58 Mallios, A. *et al.* (2011) *Navigating and Mapping with the SPARUS AUV in a Natural and Unstructured Underwater Environment,* IEEE

59 Mallios, A. *et al.* (2014) Scan matching SLAM in underwater environments. *Auton. Robots* 36, 181–198

60 Hover, E. *et al.* (2012) Advanced perception, navigation and planning for autonomous in-water ship hull inspection. *Int. J. Robot. Res.* 31, 1445–1464

61 Walter, M. *et al.* (2008) SLAM for ship hull inspection using exactly sparse extended information filters. In *IEEE International Conference on Robotics and Automation, 2008*, IEEE

62 Das, J. *et al.* (2013) Hierarchical probabilistic regression for AUV-based adaptive sampling of marine phenomena. In *2013 IEEE International Conference on Robotics and Automation*, IEEE

63 Grothues, T.M. *et al.* (2010) *Collecting, Interpreting, and Merging Fish Telemetry Data from an AUV: Remote Sensing from an Already Remote Platform*, IEEE

64 Ridao, P. *et al.* (2010) Advancing underwater robotics. *Water Power Dam Constr.* 62, 40–43

65 Smith, D. and Dunbabin, M. (2007) Automated counting of the Northern Pacific sea star in18

**Box 1. Ground robots**

Ground robots offer great potential for improving current manual surveillance methods [4] to prevent threat outbreaks, particularly in plants and animals. There are two primary domains for biosecurity applications of ground robots: agriculture and the natural environment. The constraints and problems of these two domains are discussed separately below.

*Agriculture*

In the agricultural domain, biosecurity issues are mainly considered from the perspective of optimizing production. For example, the problem of weeds [12,32,40,41] is viewed as one of weed management and not complete prevention or eradication. Both plant and animal pests can negatively impact crop output and need to be managed throughout the production cycle to minimize their impact on the final crop output. A farm offers a relatively structured physical environment from the viewpoint of robot navigation and robots are starting to be used for both detection and production tasks (Figure I). Several companies offer plant-thinning robots commercially; for example, Blue River Technologies (CA, USA) and Vision Robotics (CA, USA). Weeding robots are also available commercially; for example, ecoRobotics (Switzerland), the Robovator from F. Poulsen Engineering (Denmark), and Naio Technologies (France). Several universities across the USA [for example, Carnegie Mellon University (CMU)], Europe, and Australia (for example, the University of Sydney) are also actively developing ground robots for agricultural applications [42–45].

*Environment*

To date, ground robots have not been used for biosecurity applications in the general environment. This is because the environment can vary greatly, from being suitable for wheeled robots [for example, flat terrain along roads and tracks with good Global Positioning System (GPS) access] to being accessible only by legged robots (for example, forests and other difficult terrains with little or no GPS access).

Ground robots have the capacity to carry relatively large payloads compared with aerial and underwater robots. This enables the robots to carry more sophisticated sensor payloads and significant computing power on board and this in turn enables the robots to autonomously and adaptively monitor their environment at the ground level and report anomalies. This surveillance can significantly improve risk assessments by providing extensive and continuous sensor data.

The area of field robotics offers promising capabilities for the use of robots in biosecurity applications in the natural environment and, conversely, biosecurity applications could help to push progress in field robotics.

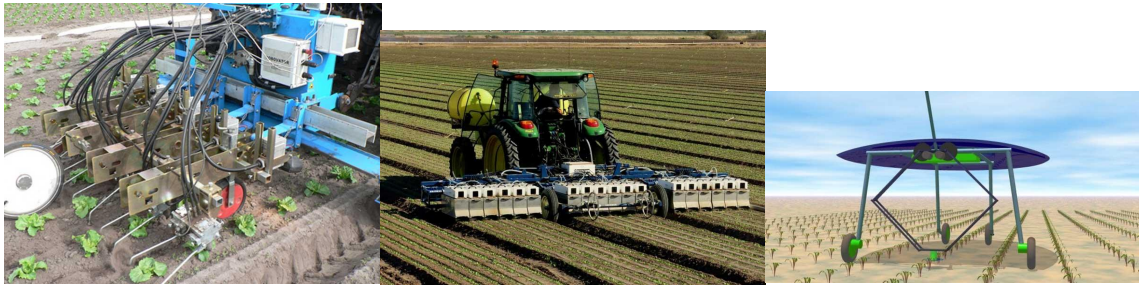

**Figure I.** Ground robots and biosecurity. Left to right: The Robovator for weed detection and eradication; a lettuce-thinning robot from Blue River Technologies; and a concept weed detection/eradication robot from ecoRobotix.

**Box 2. Aerial robots**

UAVs can fill the observation gap between remote sensing satellite systems and *in situ* observation platforms. They can also provide surveys of inaccessible or rough terrain. UAVs can target biosecurity applications that can be observed only from above (e.g., detection of invasive weeds in a rainforest [13]) or that require sampling of the aerial environment (e.g., airborne disease spores [39]).

The primary types of UAV are fixed wing, rotary wing, and lighter than air (LTA). LTA autonomous airships have potentially the longest mission times and can hover for extended periods of time over an area of interest; however, they move at slower speeds than airplanes or helicopters [46]. Fixed-wing UAVs tend to be the fastest alternative and have potentially the longest mission range but can neither hover nor operate very close to the surface. Helicopters, due to their high controllability, can be used for precise nap-of-the-earth flying as well as hovering but have shorter mission times and ranges [47]. UAV platforms for civilian use cost typically one to two orders of magnitude less than manned aircraft in their category and are also much cheaper to operate. As a result, UAVs are being deployed in increasing numbers around the world for a growing number of applications.

UAVs operate over a spatial scale of the order of $1-10^3$ km in range, with spatial observations at a ground resolution of centimeters to tens of meters depending on the flight altitude and the sensor payload. All three types of UAV can perform surveys of areas of interest at regular time intervals (typically minutes to hours) and airships and rotary-wing craft can hover over a given area for longer periods of time to allow persistent temporal monitoring of a process of interest.

Sensor payloads on UAVs for biosecurity surveys can include: red–green–blue (RGB), thermal, multispectral, and/or hyperspectral imaging sensors; lidars and/or radars for characterization of the 3D environment below the aircraft; environmental and weather sensors; and a combined GPS and inertial navigation system (INS) unit for accurate localization of sensor observations. High-value applications of UAVs in biosecurity surveys (Figure I) include: medium- and low-altitude monitoring of forests, crops, and orchards for invasive species, pests, weeds, and diseases, many of which can be detected with appropriate imagers [13]; limnological and near-shore ocean surveys for harmful algal blooms (HABs); and land surveys of wildlife for detection of invasive and feral species.

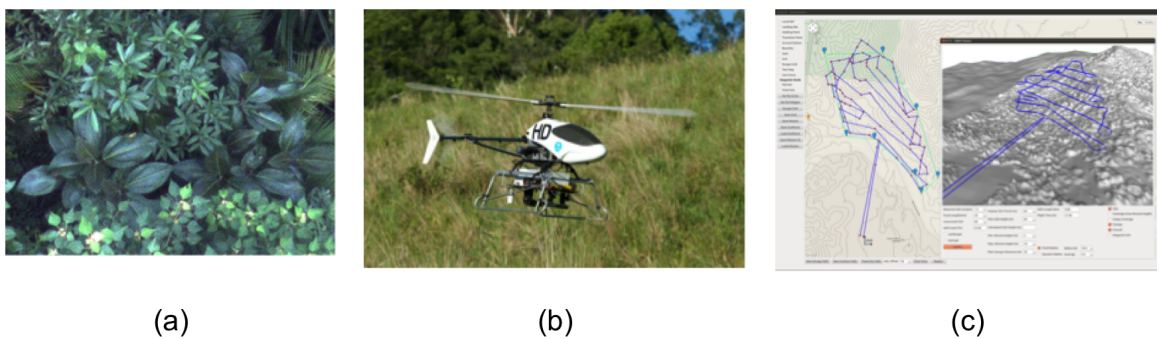

(a) (b) (c)

**Figure I.** Use of an unmanned aerial vehicle (UAV) for detection of *Miconia calvescens*. *Miconia* is an invasive plant species that is damaging rainforests (**A**) in Queensland and there is a governmental eradication program in place. (**B**) A rotary-wing UAV from the Autonomous Systems Program of the Commonwealth Scientific and Industrial Research Organisation (CSIRO) is used to conduct aerial surveys in rough terrain and flying close to the canopy of the forest. (**C**) An example of the survey pattern planned automatically by the supervisory system and executed by the UAV.

**Box 3. Aquatic robots**

Aquatic robots comprise AUVs, remotely operated vehicles (ROVs), and ASVs. They can be instrumental in detecting incursions of invasive species into protected areas, either organically over time or through attachment to the hulls of ships that travel from other areas where the species is prevalent.

Aquatic robots can be deployed for standalone tasks such as ship hull inspection, cleaning [48–50], reef monitoring by the Commonwealth Scientific and Industrial Research Organisation (CSIRO) Starbug (Figure IA), or sample collecting by the Monterey Bay Aquarium Research Institute (MBARI) Gulper (Figure IB) or be used alongside static sensor networks such as data mules and mobile nodes with high-value sensor payloads [51,52]. A CSIRO ASV (Figure IC) can be used as a mobile node for water quality monitoring to detect HABs in a water storage reservoir connected to a larger, floating static sensor network.

Aquatic robots can carry underwater vision and sonar sensors to detect and characterize submerged biosecurity threats. Although most shallow water biosecurity surveillance is currently being done by divers, robotic solutions based on ROVs are now emerging for submerged port inspection and in-water ship hull inspection [49] and cleaning [50] applications, to prevent the introduction of invasive species [2]. ROVs can operate at almost any depth, perform high-resolution surveys of an area of interest and perform interventional tasks. However, these require permanent connection to a surface vessel or a larger AUV by a tether, constraining their maximum operation range, as in the case of the RovingBat (Figure ID).

---

[2] Department of Fisheries, Western Australia (2014) *Asian Green Mussel Factsheet* (http://www.fish.wa.gov.au/Documents/biosecurity/marine_pest_fact_sheet_asian_green_mussel.pdf)

AUVs are more suitable for tracking biosecurity events as they can cover a spatial scale of the order of 1–20 km$^2$ in range with a resolution of few centimeters and do not require tethering to another vessel, allowing them to freely maneuver during a mission. They can explore the seafloor with no gaps in deep [53–56] or shallow water [57,58], to inspect submerged structures [59,60] or ship hulls [60,61], take samples in HABs [62], or perform fish monitoring [63] or pest population control [64,65]. AUV mission duration is usually constrained by their endurance. Glider-type AUVs overcome this problem by using passive propulsion systems, significantly increasing mission ranges and lifetimes [66–70].

ASVs cover areas of several square kilometers with high-resolution scanning. However, they require complex navigation systems that follow maritime rules and regulations while taking into account the traffic around them. Moreover, their sensors are inherently constrained to operate close to the surface, limiting the depth and resolution of seabed inspections.

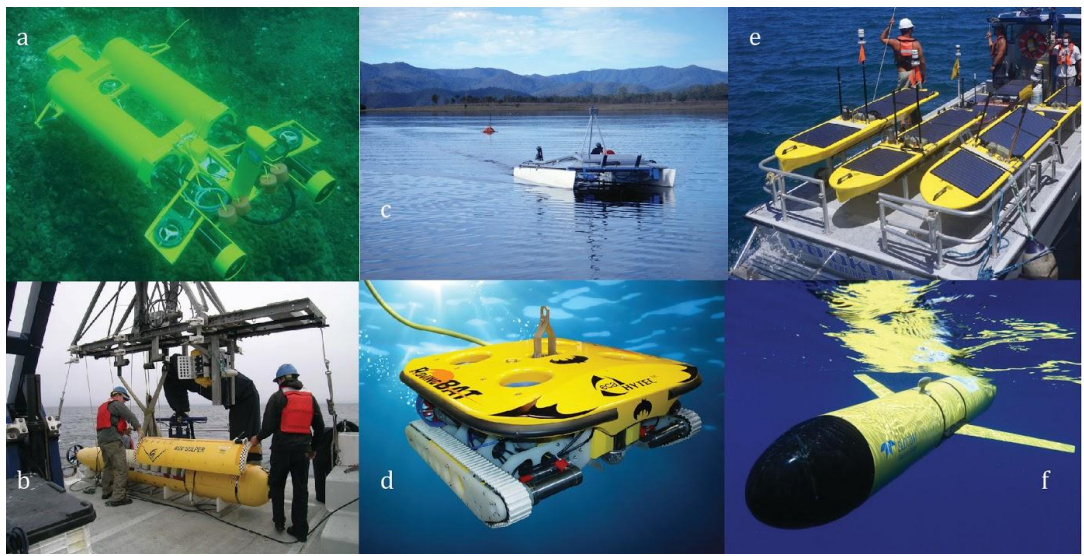

**Figure I.** Aquatic robots for biosecurity. (**A**) Starbug autonomous underwater vehicle (AUV) from the Commonwealth Scientific and Industrial Research Organisation CSIRO) used for counting crown-of-thorns starfish. (**B**) A Gulper AUV operated by the Monterey Bay Aquarium Research Institute (MBARI) capable of detecting harmful algal blooms (HABs) (http://www.mbari.org). (**C**) An autonomous surface vehicle from CSIRO used for water quality monitoring (http://www.csiro.au). (**D**) The RovingBat remotely operated vehicle (ROV) from ECA Robotics used for ship hull inspection and cleaning (http://www.eca-robotics.com). (**E**) Wave Glider SV3s from Liquid Robotics, used for persistent monitoring of algal blooms (http://liquidr.com). (**F**) A Slocum glider AUV developed by the Woods Hole Oceanographic Institution (WHOI) to monitor phytoplankton dynamics and bottom water oxygen during HABs (http://www.whoi.edu).